\documentclass[letterpaper]{article}

\usepackage{aaai16}
\usepackage{times}
\usepackage{helvet}
\usepackage{courier}
\usepackage{epstopdf}
\usepackage{amsmath,amsfonts}
\usepackage{multicol,multirow}
\usepackage{amsthm}
\usepackage{algorithm}
\usepackage[noend]{algpseudocode}
\usepackage{graphicx,subfigure,epsfig}
\usepackage{url}
\usepackage{color}
\usepackage{array}
\usepackage{diagbox}

\newcommand{\secref}[1]{Section \ref{#1}}
\newcommand{\figref}[1]{Figure \ref{#1}}
\newcommand{\eqnref}[1]{Eq. (\ref{#1})}
\newcommand{\tabref}[1]{Table \ref{#1}}

\newcommand{\cut}[1]{}
\newcommand{\pair}[2]{$\langle$#1, #2$\rangle$}

\newcolumntype{I}{!{\vrule width 1pt}}
\newlength\savedwidth
\newcommand\whline{\noalign{\global\savedwidth\arrayrulewidth
                            \global\arrayrulewidth 1pt}%
                   \hline
                   \noalign{\global\arrayrulewidth\savedwidth}}
\newcommand{\tabincell}[2]{\begin{tabular}{@{}#1@{}}#2\end{tabular}}

\title{Representing Verbs as Argument Concepts}

\author{
Yu Gong{\small $~^{1}$} \and Kaiqi Zhao{\small $~^{2}$} \and
Kenny Q. Zhu{\small $~^{3}$}\\
Shanghai Jiao Tong University, Shanghai, China\\
{\small $~^{1}$}gy910210@163.com, {\small $~^{2}$}kaiqi\_zhao@163.com,
{\small $~^{3}$}kzhu@cs.sjtu.edu.cn}

\begin{document}
\maketitle

\begin{abstract}
Verbs play an important role in the understanding of natural
language text. This paper studies the problem of abstracting
the subject and object arguments of a verb into a set of noun
concepts, known as the ``argument concepts''.
This set of concepts, whose size is parameterized,
represents the fine-grained semantics of a verb. For example,
the object of ``enjoy'' can be abstracted into time, hobby and event, etc.
We present a novel framework to automatically infer human
readable and machine computable action concepts
with high accuracy.
\end{abstract}

\section{Introduction}
Verb plays the central role in both syntax and semantics of natural language sentences.
The distributional hypothesis~\cite{harris1954distributional,miller1991contextual}
shows that it is possible to represent the meaning of a word by the distributional
properties of its context, e.g., its surrounding words in a window.
A verb has a unique role in a sentence because it maintains dependency relation
with its syntactic arguments such as the subject and the object.
Therefore, it is possible to use the
distribution of immediate arguments of a verb
to represent its meaning, such as ReVerb~\cite{fader2011identifying}.
Such an approach is a form of
``bag-of-words'' (BoW) approach. The common criticisms of the BoW approach
are i) perceived orthorgonality of all words despite some of them sharing
similar or related meanings; ii) its high dimensionality and high
cost of computation; and
iii) poor readibility to humans.

To ameliorate these limitations, a natural solution is to represent the
arguments by their abstract types, rather than the words themselves.
It is reasonable to assume that a verb represents different meanings,
or different senses, if it's used with different types of arguments.
To that end, FrameNet~\cite{baker1998berkeley} and VerbNet~\cite{kipper2000class}
are examples of human-annotated lexicons that include verbs and their
meanings (called {\em frames}) and the different types of
their arguments (called {\em thematic roles} or {\em semantic roles}).
Due to the excessive cost of constructing such
lexicons, as well as their intentional shallow semantic nature,
the abstraction of verb arguments is very coarse-grained. For example,
in FrameNet, the verb ``eat'' has just one frame, namely ``Ingestion'',
and its direct object has just one role, ``Ingestibles''.
Furthermore, the lexical coverage of these
resources are very limited. FrameNet, which is the most popular and best
maintained among the three, consists of just 3000 verbs and 1200
frames.

The BoW approach is too fine-grained while the semantic role approach
is too coarse-grained.
In this paper, we seek to strike a balance between these two
extremes. Our goal is to automatically infer a tunable set of
human-readable and machine-computable abstract concepts for
the immediate arguments~\footnote{We only consider subjects and
direct objects in this paper, though other arguments may be inferred
as well.} of each verb from a large text corpus.
By ``tunable'', we mean that the granularity of
the concepts can be parameterized by the size of the set to be returned.
The larger the set, the finer-grained the semantics.
The vocabulary of the concepts
comes from an existing taxonomy of concepts or terms such as
Probase~\cite{wu2012probase}  or WordNet~\cite{miller1998wordnet}.
For instance, the direct object of verb ``eat'' may be conceptualized
into ``food'', ``plant'' and ``animal''.

One potential solution toward this goal is selectional preference
(SP), originally proposed by Resnik\shortcite{resnik1996selectional}.
Class-based SP computes whether a class of terms is a preferred argument
to a verb. Together with a taxonomy of concepts,
SP can produce a ranked list of classes that are the most appropriate
subjects or objects of a verb. However, for the purpose of representing verbs,
SP has the following drawback: it doesn't allow the granularity of the concepts
to be tuned because it computes a selectional preference score between the
verb and {\em every} possible concept in the taxonomy. The top $k$ concepts
do not necessarily cover all the aspects of that verb because these
concepts may semantically overlap each other.
Clustering-based SP and LDA-based SP~\cite{ritter2010latent}
find tunable classes with low overlaps, but the classes are either
word clusters or probabilistic distributions of words,
which are not abstracted into concepts. Without
associating the classes to concepts in taxonomies,
the model loses the ability of generalization.
For example, if ``eat McDonalds'' does not appear in the training
data, clustering- and LDA-based SP cannot recognize ``McDonalds''
as a valid argument to ``eat'', since ``McDonalds'' is not
a member of any inferred clusters or word distributions.

In this paper, we first introduce the notion of taxonomy (\secref{sec:tax})
and define the argument conceptualization problem,
which asks for $k$ concepts drawn from a taxonomy
that generalize as many possible arguments
of a verb as possible, and with bounded overlap with each other
(\secref{sec:problem}). We present the system
to generate tunable argument concepts through a branch-and-bound
algorithm (\secref{sec:algo}) and show in experiments that our system can
generate high quality human-readable and machine-computable
argument concepts (\secref{sec:eval}). Some related work will be discussed
(\secref{sec:related}) before we draw some concluding remarks
(\secref{sec:conclude}).

\section{Taxonomy}
\label{sec:tax}
We use a taxonomy as the external classification knowledge for
conceptualizing the arguments of verbs.
A taxonomy is a directed graph $(V, E)$,
Here, $V$ is a set of terms, $E$ is a set of binary ``isA'' relations
\[E=\{(e,c)| e\in V, c\in V, e~ isA~ c\},\]
where $e$ is called an {\em entity},
$c$ is called a {\em concept}, and $c$ is said to {\em cover} $e$.
Most terms in $V$ are
both concepts and entities; terms with zero outdegree in the graph are entities only.
In this paper, we consider two different taxonomies,
namely {\em WordNet}~\cite{miller1998wordnet} and {\em Probase}~\cite{wu2012probase}.
WordNet organizes words into sets of synonyms
(called {\em synsets}) along with ``isA'' relation between two synsets.
Each word may belong to multiple synsets and have multiple
hypernyms (concepts) or hyponyms (entities).
Probase covers a lot of named entities and multi-word expressions
(e.g., Microsoft, Star Wars) which may not be covered by WordNet.
This feature allows us to extract more precise arguments.

\section{Problem Formulation}
\label{sec:problem}
We begin with an informal definition of the
{\em argument conceptualization} problem.
Given a collection of argument instances of the same argument
type (e.g., object or subject) of a verb,
we want to pick $k$ concepts from the taxonomy
that subsume as many instances as possible.
We would also like these $k$ concepts to
have little overlap with each other.
The intuition is that each of the $k$ selected concepts represents a unique
sense with small semantics overlap and the $k$ concepts collectively cover the majority
uses of that verb.

We define semantics {\em overlap} between two concepts as:
$$Overlap(c_1,c_2)=\frac{|E_{c_1}\cap E_{c_2}|}{min\{ |E_{c_1}|,|E_{c_2}| \}},$$
where $E_c$ is the set of all entities covered by concept $c$ in
the taxonomy.

Then, we formulate the argument conceptualization problem as
a problem of finding maximum weighted $k$-cliques. Consider a \emph{concept graph}
$G=(C,L,W)$, which has a collection of concepts $C$ in a taxonomy,
and a set of edges $L$ in which each edge connects two concepts that
have an overlap less than a predefined threshold $\tau$. $W$ stands for
weights for the concepts in the graph.
Each weight intuitively represents the quality of
the concept with respect to the verb.

\begin{figure}[th]
\centering
\includegraphics[width=0.8\columnwidth]{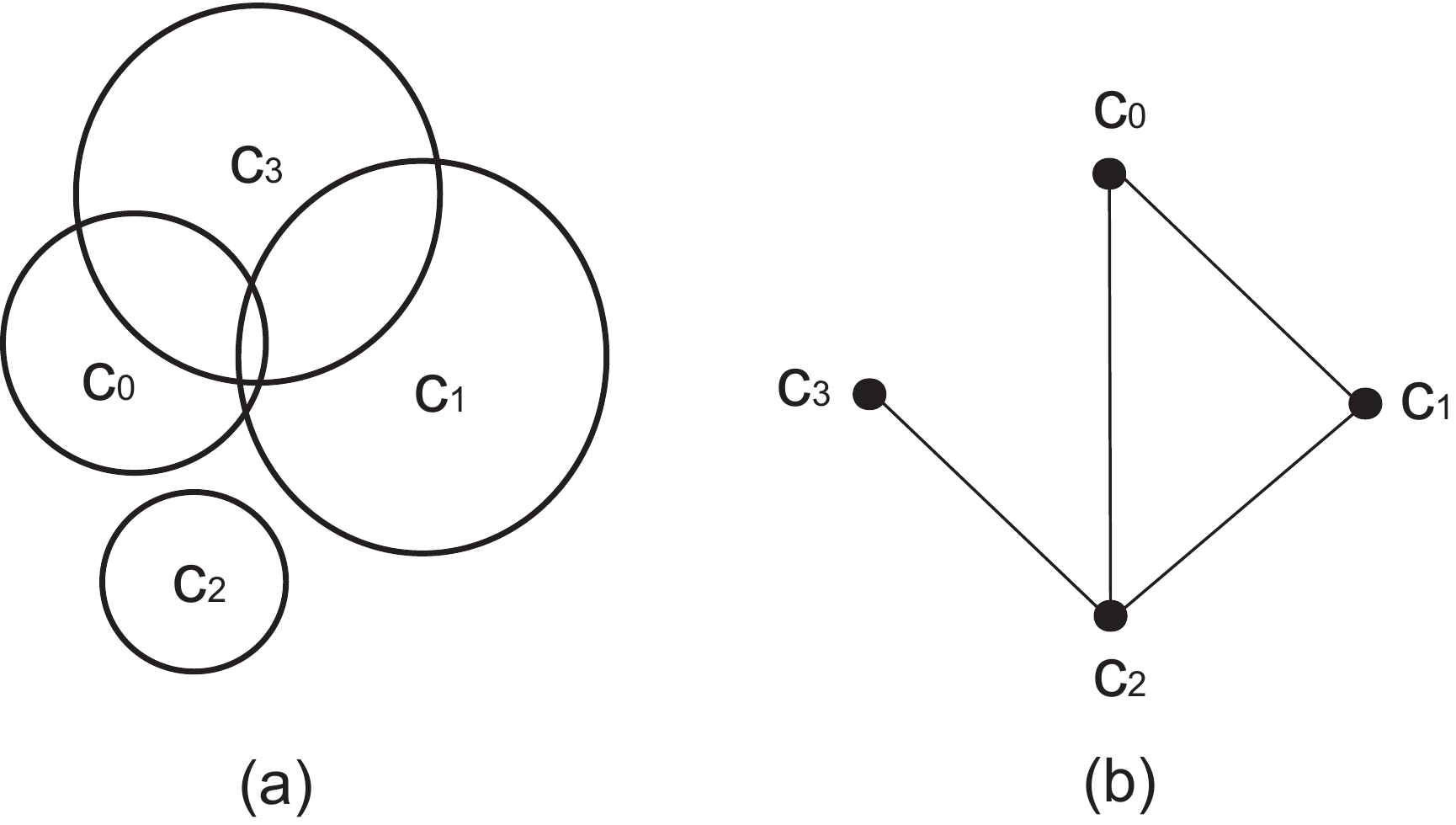}
\caption{(a) 4 concepts in the entity space
(b) corresponding concept graph}
\label{fig:graph_model}
\end{figure}

\figref{fig:graph_model} shows 4 concepts in an illustrative 2-dimensional
entity space (a), as well as their corresponding concept graph (b).
Each circle $c_i$ in (a) represents a set of entities covered by concept $c_i$.
Because the overlap between $c_0$ and $c_3$ and between $c_1$ and $c_3$
is high  ($>\tau$), (b) is a fully connected graph (clique) minus only
two edges: $l_{c_0,c_3}$ and $l_{c_1, c_3}$.

The argument conceptualization problem is then transformed to
finding the $k$-clique with maximum combined weight.

A straightforward way to define the weight for each concept is
counting the number of argument instances it subsumes according to the
isA taxonomy (used as baseline method in \secref{sec:eval}).
This assumes that all argument instances of a verb are of
equal importance, which is not true in practice.
We thus generalize the importance of an argument $e$ to a verb $v$
by a quality function $Q_v(e)$, which we will discuss in detail
in \secref{sec:qe}.
Consequently, the weight of concept $c$ for verb $v$ is
defined as
\begin{equation}
w_v(c)=\sum_{e\in \{e|e\;\text{isA}\;c\}}{Q_v(e)}.
\end{equation}
The argument conceptualization problem is to find
a $k$-clique (which forms a concept set as $C_k$)
in the graph $G$ which maximizes
\begin{equation}
\label{eq:f}
f_v(C_k)=\sum_{c\in C_k}{w_v(c)}.
\end{equation}

We parameterize the number ($k$) of argument concepts of a
verb because i) different verbs have different number of senses; and ii)
even for the same verb, there is no agreement on the exact number of its
senses because one meaning can always be divided into a number of
finer-grain meanings.
For example, in Oxford English Dictionary~\cite{oxford},
the transitive verb ``eat'' has 4 senses (or definitions),
while in Cambridge Dictionary~\cite{cambridge} it has just one meaning.

\section{Framework}
\label{sec:algo}
Our framework consists of three main steps:
argument extraction, argument weight computation and
argument conceptualization.
In the argument extraction component, we extract the arguments
of the verb from a dependency parsed sentence by several
dependency relations (``nsubj'', ``agent''
for subject extraction and ``nsubjpass'', ``dobj'' for object extraction).
In the argument weight computation component,
we pre-compute the weight for each argument instance
(see \secref{sec:qe}). In the argument conceptualization, we
build the concept graph and use a branch-and-bound algorithm
(see \secref{sec:bb})
to solve the argument conceptualization problem.

\subsection{Argument Weight Computation}
\label{sec:qe}
Since many of the existing dependency parser systems are noisy~\cite{manning2014stanford}.
Our observations showed that some errors follow certain patterns.
For example, ``food'' in ``food to eat''
is usually incorrectly labeled as the subject of ``eat'',
and the same goes for ``water to drink'', ``game to play'', etc.
Similarly, ``time'' in ``play this time'' and ``play next time''
is incorrectly labeled as the object of ``play''.
We also discovered that if an argument is incorrect due to parsing,
it is often extracted from just a couple of patterns. Conversely,
if an argument is correct for the verb, it probably appears under
many different patterns.
Consider ``corn'' as an object of verb ``eat''.
It appears in 142 patterns, e.g., ``eat corn'', ``eat expensive corn'', ``eat not only corn'',
etc., each of which gives a different dependency structure.  However,
``habit'' only appears in 42 patterns like ``eating habit''.
We follow this observation and assume that correct arguments
generally are likely to appear in more patterns than the wrong ones.
We define a pattern as a subtree in the dependency tree
according to two rules:
\begin{itemize}
\item The argument and one of its children
form a pattern:
$$\{POS_{arg}, DEP_{arg}, POS_{child}, DEP_{child}\},$$
where $POS$ and $DEP$ stand for POS tag and dependency type, respectively.
\item The argument and its siblings form another pattern:
$$\{POS_{arg}, DEP_{arg}, POS_{sib}, DEP_{sib}\}.$$
\end{itemize}

For each argument $e$ of verb $v$, we collect the set of its patterns
$M_{e,v}$, and
use the entropy to measure the correctness, where a higher
entropy value means that the argument is more informative w.r.t.
the patterns, and hence more likely to be a valid argument.
The entropy is defined as:
\begin{equation}
\text{Entropy}_v(e)=-\sum_{m\in M_{e,v}}{P(m)\log{P(m)}}
\end{equation}

Moreover, even if an argument is valid under a verb, it may be
less relevant. For example, while ``fruit'' is highly relevant to ``eat'',
``thing'' is not because it can be the object of many other verbs.
To this end, we use a binary version of mutual information to
measure the relatedness between two terms.
The mutual information $MI_v(e)$ is defined as:
\begin{equation}
MI_v(e)=
\begin{cases}
1 & \mbox{if}~\ p(v,e)\log \frac{p(v,e)}{p(v)p(e)}> 0,\\
-1 & \rm{otherwise}.
\end{cases}
\end{equation}

In essence, the entropy measures the correctness of the argument, while
mutual information measures its correlation with the verb.
We compute the quality of an argument by combining these two measures:
\begin{equation}
Q_v(e)=\text{Entropy}_v(e)\times \text{MI}_v(e).
\label{eq:qe}
\end{equation}

\subsection{A Branch-and-Bound Algorithm}
\label{sec:bb}
\begin{figure}[th]
\centering
\includegraphics[width=0.6\columnwidth]{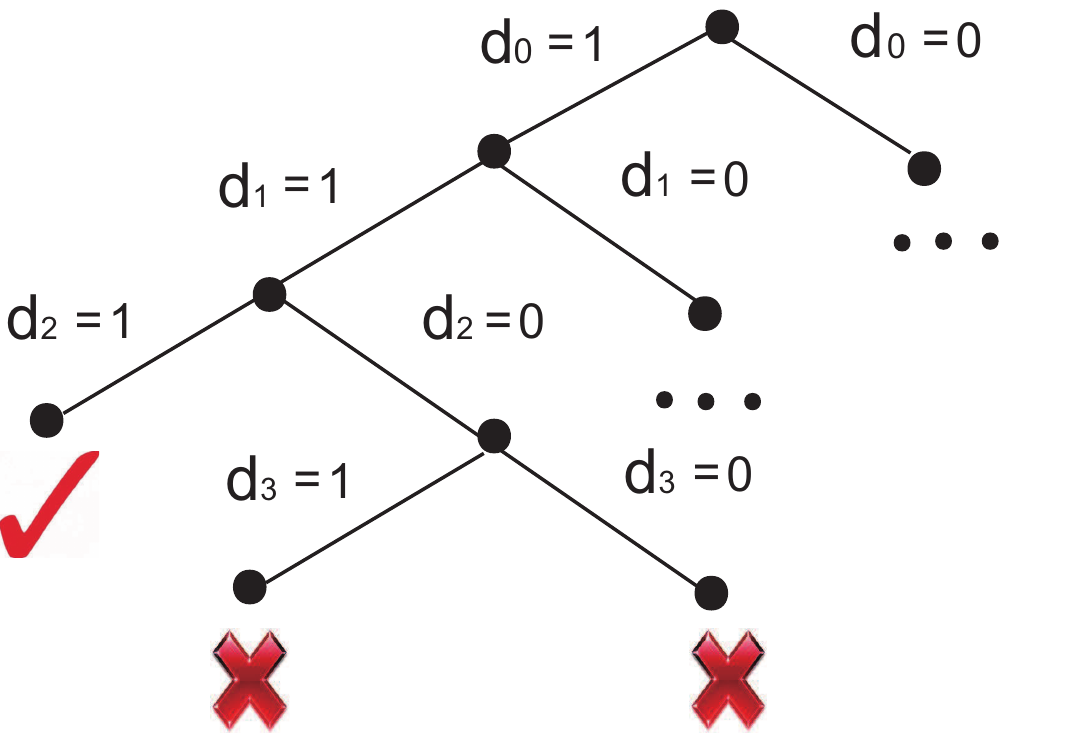}
\caption{A Snapshot of the Binary Decision Tree with $k=3$}
\label{fig:search_tree}
\end{figure}

Because the concept space in a general-purpose taxonomy is large,
we propose a branch-and-bound algorithm to efficiently search for the solution.
The details of our algorithm are shown in Algorithm \ref{al:backtrack}.
We model each solution as a binary vector of size $|C|$ ($C$ is
the set of all concepts in the taxonomy) in which exactly
$k$ elements of the vector are set to 1 while others are set to 0.
The search space is represented by a binary decision tree
where the nodes at each level indicate the decision to include a concept in
the solution or not. The complete search space contains $2^{|C|}$ nodes.
Take the concept graph in \figref{fig:graph_model} as an example.
The corresponding search space is shown in \figref{fig:search_tree},
in which $d_i=1$ means to include $c_i$ in the solution, and $d_i=0$ means
otherwise. For $k=3$, the concept set $\{c_0,c_1,c_2\}$ is
a valid solution, which is marked by the path
$(d_0=1) \rightarrow (d_1=1) \rightarrow (d_2=1)$. The key insight in this algorithm
is that, even though the search space is exponential, a subtree can be
pruned if its path from the root already contains a valid solution, or
if the current path doesn't have the potential to produce a better solution
than the current best.

\begin{algorithm}[th]
\small
\caption{Argument Conceptualization}
\label{al:backtrack}
\begin{algorithmic}[1]
\Function{AC}{$W, C, L, k$}
\State $\{c_0,...,c_{|C|-1}\}\leftarrow$ Sort concepts $c\in C$ in\ the\ descending\ order\ of $w_v(c)$.
\State $\pi_{max} \leftarrow 0,\pi_{c} \leftarrow 0,ck \leftarrow 0$
\State $d_{max}\leftarrow\{0,...,0\},d\leftarrow\{0,...,0\}$
\State BB($0$)
\If{$ck = k$}
\State \textbf{return} $d_{max}$
\Else
\State \textbf{No solution}
\EndIf
\EndFunction
\Statex
\Function{BB}{$i$}
\If{$i\geq |C|$}
\State \textbf{return}
\EndIf
\If{$ck = k$}
\If{$\pi_{c}>\pi_{max}$}
\State $\pi_{max} \leftarrow \pi_{c}, d_{max} \leftarrow d$
\EndIf
\State \textbf{return}
\EndIf
\If{ISCLIQUE($L,i$) $= TRUE$ and BOUND($i$)$>\pi_{max}$}
\State $ck \leftarrow ck+1, \pi_{c} \leftarrow \pi_{c}+ws_v(c_i), d_i \leftarrow 1$
\State BB($i+1$)
\State $ck \leftarrow ck-1, \pi_{c} \leftarrow \pi_{c}-ws_v(c_i), d_i \leftarrow 0$
\EndIf
\If{BOUND($i+1$) $> \pi_{max}$}
\State $d_i \leftarrow 0$
\State BB($i+1$)
\EndIf
\State \textbf{return}
\EndFunction
\Statex
\Function{ISCLIQUE}{$L, i$}
\For{$j$ from $0$ to $i-1$}
\If{$d_j = 1$}
\If{$(c_i, c_j)\not\in L$ and $(c_j, c_i)\not\in L$}
\State \textbf{return} $FALSE$
\EndIf
\EndIf
\EndFor
\State \textbf{return} $TRUE$
\EndFunction
\Statex
\Function{BOUND}{$i$}
\State $b \leftarrow \pi_{c}$
\For{$j$ from $i$ to $\min\{i+k-ck-1, |C|-1\}$}
\State $b \leftarrow b+ws_v(c_{j})$
\EndFor
\State \textbf{return} $b$
\EndFunction
\end{algorithmic}
\end{algorithm}

Suppose the partial solution of the first $i$ levels in the tree
are $(d_0, d_1, ..., d_{i-1})$ and
the current best solution has a score (computed by \eqnref{eq:f}).
We use $d_{max}$ and $\pi_{max}$ to store the
best solution and its score found thus far; and use $d$ and $\pi_{c}$ to
represent the current partial solution and its partial score.
Variable $ck$ stands for the number of concepts that have been set to
1 in the current decision path, i.e.,
\[ck=\sum_{n=0}^{i-1}d_n.\]

The main function BB($i$) searches through the tree in a depth-first manner.
It returns when it reaches the leaf node (Line 11-12) or when it has found a
solution (Line 13-16). If the solution is better than the current best,
the current best solution is updated. The function traverses one
more level to include concept $c_i$ (Line 17-19) if it forms
a clique with the currently chosen concepts (ISCLIQUE function)
and if the maximum possible score with $c_i$ is better than
the current best score (BOUND function).

A crucial optimization in this algorithm is that we first sort
all concepts in $C$ in the descending order of their weighted scores (Line 2).
This allows us to quickly
compute the bound (Line 33-34) in linear time (against $k$), i.e., simply
compute the total score of the next $k-ck$ concepts down the decision
tree hierarchy, rather than sorting all the remaining concepts.

\section{Experimental Results}
\label{sec:eval}
In this section, we first show how we prepare the data for argument
conceptualization. Then, we use some example concepts
generated by our algorithm to show the advantage of our algorithm (AC)
against selectional preference (SP), FrameNet~\cite{baker1998berkeley} and ReVerb~\cite{fader2011identifying},
as well as our baseline approach (BL) which considers equal
weight for each argument (see \secref{sec:problem}).
We also quantitatively evaluate the accuracies of AC, BL and SP
on Probase. Finally, we apply our algorithm to
an NLP task known as argument
identification~\cite{gildea2002necessity,abend2009unsupervised,meza2009jointly}
and show that concepts generated by AC
achieve better accuracy against BL, SP,
Reverb and a state-of-the-art semantic role labeling tool (using FrameNet) on both taxonomies.

\subsection{Experimental Setup}
\label{sec:preprocess}
We use our algorithm to conceptualize subjects and objects
for 1770 common verbs from Google syntactic
N-gram~\cite{goldberg2013dataset,googlengram} using
Probase and WordNet as isA taxonomies.
\footnote{All evaluation data sets and results
are available at \url{http://adapt.seiee.sjtu.edu.cn/ac}.}
From 1770 verb set, we
sample 100 verbs with probability proportional to
the frequency of the verb. This set of 100 verbs (Verb-100) is
used for quantitative experiments including evaluating the
accuracy of argument concepts and the accuracy of
argument identification.

All argument instances we use in this work come
from {\em Verbargs} and {\em Triarcs} packages
of the N-gram data.
From the labeled dependency trees, we extract
subject-verb dependency pairs (nsubj, agent) and
object-verb dependency pairs (dobj, nsubjpass).
We expand the subject or object, which is a word, into
a phrase recognizable by Probase/WordNet by sliding a window
across the subtree rooted at the argument word.

For the system parameters, we set the maximum overlap
threshold between two concepts to 0.2, and the number of concepts
$k$ to $\{5, 10, 15\}$ to evaluate argument concepts of
different granularity.
In practice, the number $k$ can be set differently for different verbs,
which we view as an advantage of the framework.

\subsection{Conceptualization Results}
We compare the concepts learned by AC with the concepts learned by
BL, FrameNet elements, Reverb arguments, and concepts learned by SP.
ReVerb is an open information extraction system
that discovers binary relations\footnote{ReVerb extracts general relations
instead of verb predicates, e.g., XXX heavily depends on YYY.} from the web
without using any predefined lexicon. ReVerb data contains
15 million subject-predicate-object triple instances without
any abstraction or generalization.

\begin{table*}[ht]
  \centering
  \small
  \caption{Example subject/object concepts from 5 lexicons}
    \begin{tabular}{cIrIl|l|l|l|l}
    \whline
    Verb  &       & \multicolumn{1}{c}{AC Concepts} & \multicolumn{1}{|c}{BL Concepts} & \multicolumn{1}{|c}{FrameNet} & \multicolumn{1}{|c}{ReVerb} & \multicolumn{1}{|c}{SP Concepts} \\
    \whline
    \multirow{2}[6]{*}{accept} & Subj  & \tabincell{l}{person, community, \\ institution, player,\\ company} & \tabincell{l}{topic, name, \\ group, feature, \\ product} & \tabincell{l}{Recipient, \\ Speaker, \\ Interlocutor}  &  \tabincell{l}{Student, an article, \\ the paper, Web browser, \\ Applications} & \tabincell{l}{world, group, \\ person, term, \\ safe payment method}   \\
          \cline{2-7}
          & Obj   & \tabincell{l}{document, payment, \\ practice, doctrine, \\ theory} & \tabincell{l}{factor, feature, \\ product, activity, \\ person} & \tabincell{l}{Theme, \\ Proposal}  & \tabincell{l}{the program, publication, \\ HTTP cookie, the year, \\ credit card}  & \tabincell{l}{topic, concept, \\ matter, issue, \\ word}   \\
    \hline

    \multirow{2}[6]{*}{enjoy} & Subj  & \tabincell{l}{group, community, \\ name, country, \\ sector} & \tabincell{l}{name, topic, \\ group, feature, \\ product} & Experiencer  &  \tabincell{l}{people, ive, Guests, \\ everyone, someone} & \tabincell{l}{world, stakeholder, \\ group, person, \\ actor}   \\
          \cline{2-7}
          & Obj   & \tabincell{l}{benefit, time, hobby, \\ social event, \\ attraction} & \tabincell{l}{factor, activity, \\ feature, product, \\ person} & Stimulus  &  \tabincell{l}{life, Blog, Breakfirst, \\ their weekend, a drink}  & \tabincell{l}{benefit, issue, \\ advantage, factor, \\ quality}   \\
    \hline

    \multirow{2}[6]{*}{submit} & Subj  & \tabincell{l}{group, community, \\ name, term, \\ source} & \tabincell{l}{topic, name, \\ group, feature, \\ product} & Authority  &  \tabincell{l}{no reviews, Project, \\ other destinations, \\ HTML, COMMENTS} & \tabincell{l}{large number, number, \\ stakeholder, position, \\ group}   \\
          \cline{2-7}
          & Obj   & \tabincell{l}{document, format, \\ task, procedure, \\ law} & \tabincell{l}{factor, feature, \\ activity, product, \\ person} & Documents  & \tabincell{l}{one, review, \\ a profile, text, \\ your visit dates}  & \tabincell{l}{document, esi online tool, \\ material, nickname, \\ first name}   \\
    \whline
    \end{tabular}
  \label{tab:results}
\end{table*}

Table \ref{tab:results} shows 3 example verbs and their argument concepts (AC \& BL),
FrameNet semantic roles (FN), ReVerb argument instances (RV) as well as
selectional preference (SP) concepts for the verbs' subjects and objects.
The number of concepts $k$ is set to 5 for AC \& BL, and the top 5 instances/concepts are showed for
RV and SP.
We can observe that the semantic roles in FN are too general, while RV instances
are too specific. Both inevitably lose information:
FN is a manually constructed lexicon by experts thus
cannot scale up well, while ReVerb is automatically extracted from massive
English sentences and hence comes with abundant errors (e.g., {\em ive}
as a subject of ``enjoy'').
SP does not consider semantic overlaps between argument concepts.
BL assumes that all argument instances of a verb are of
equal importance, which is not true in practice. It tends to generate uninformative concepts such as ``factor'' and ``feature''.
Compared to the other methods, AC generates concepts with
tunable granularity and low semantic overlap. These concepts are more comprehensive
and more accurate.

To quantitatively compare our algorithm to BL and SP,
we ask three native English speakers to
annotate whether the concepts generated by AC, BL and SP are
the correct abstraction of the verb's arguments.
The majority votes are used as the ground truth. We compute the percentage of
correct concepts as accuracy, and report the accuracy of
AC, BL and SP in \tabref{tab:precision}.
AC generates more accurate concepts than BL and SP mainly because AC
considers the quality of argument instances extracted from dependency
and the semantic overlap between concepts.
BL performs worse than SP because the noise caused by parsing
error is not considered, but SP considers the association between the verb and
arguments which implicitly gives a low rank to the incorrect arguments.

\begin{table}[ht]
\centering
\scriptsize
\caption{Accuracy of AC, BL and SP concepts}
\begin{tabular}{cIc|c|c|c|c|c}
\whline
\multirow{2}{*}{k} & \multicolumn{3}{c|}{Subject} & \multicolumn{3}{c}{Object}\\
\cline{2-7}
& AC & BL & SP &  AC & BL & SP \\
\whline
5 &\bf 0.88 & 0.49 & 0.58 &\bf 0.97 & 0.63 & 0.62 \\
\hline
10 &\bf 0.86 & 0.47 & 0.56 &\bf 0.94 & 0.61 & 0.65 \\
\hline
15 &\bf 0.85 & 0.43 & 0.58 &\bf 0.91 & 0.60 & 0.66 \\
\whline
\end{tabular}
\label{tab:precision}
\vspace{-1.5em}
\end{table}

\subsection{Argument Identification}
In the argument identification task, we use the inferred argument concepts
to examine whether a term is a correct argument to a verb in a sentence.
To evaluate the accuracy of argument identification, for each verb in Verb-100,
we first extract and randomly select 100 sentences containing the verb from the Engish Wikipedia
corpus. We then extract \pair{verb}{obj} and \pair{verb}{subj} pairs from these 10,000 sentences.
Apart from parsing errors, most of these pairs are correct
because Wikipedia articles are of relatively high quality.
We roughly swap the subjects/objects from half of these pairs with
the subject/object of a different verb, effectively creating
incorrect pairs as negative examples.
For example, if we exchange ``clothing'' in ``wear clothing'' with the ``piano''
in ``play piano'', we get two negative examples ``wear piano''
and ``play clothing''.
Finally, we manually label each of the 20,000 pairs to be correct or not,
in the context of the original sentences.  As a result, we have a test
set of 10,000 \pair{verb}{obj} and \pair{verb}{subj} pairs in which
roughly 50\% are positive and the rest are negative.

We compare AC with BL, SP, ReVerb and Semantic Role Labeling (SRL) as follows:
\begin{itemize}
\item {\bf AC \& BL \& SP}: Check if the test term belongs to any of the $k$ argument concepts (isA relation) of the target verb.
\item {\bf ReVerb}: Check if the test term is contained by the
verb's list of subjects or objects in ReVerb.
\item {\bf SRL}: SRL aims at identifying the semantic arguments
of a predicate in a sentence, and classifying them into different
semantic roles.
We use ``Semafor''\cite{chen2010semafor}, a well-known SRL tool,
to label semantic arguments with FrameNet in the sentences,
and check if the test term is recognized as a semantic argument of
the target verb.
\end{itemize}

\begin{table}[th]
  \centering
  \scriptsize
  \caption{Accuracy of argument identification}
    \begin{tabular}{cIcIc|c|c|c|c|c|l@{}}
        \whline
        \multirow{2}{*}{} & \multirow{2}{*}{k} & \multicolumn{3}{c|}{Probase} & \multicolumn{3}{c|}{WordNet} & \multirow{2}{*}{\diagbox[dir=SE,height=2em,trim=rl]{RV}{SRL}} \\
        \cline{3-8}
             & & AC & BL & SP & AC & BL & SP &   \\
        \whline
            \multirow{3}{*}{Subj}
            & 5 & \bf 0.81 & 0.50 & 0.70 & 0.55 & 0.54 & 0.54 & \multirow{3}{*}{\diagbox[dir=SE,height=3em,trim=rl]{0.54}{0.48}} \\
        \cline{2-8}
            & 10 & \bf 0.78 & 0.50 & 0.72 & 0.57 & 0.54 & 0.55 &  \\
        \cline{2-8}
            & 15 & \bf 0.77 & 0.49 & 0.72 & 0.58 & 0.54 & 0.56 &  \\
        \whline
            \multirow{3}{*}{Obj}
            & 5 & \bf 0.62 & 0.51 & 0.58 & 0.50 & 0.46 & 0.50 & \multirow{3}{*}{\diagbox[dir=SE,height=3em,trim=rl]{0.47}{0.50}} \\
        \cline{2-8}
            & 10 & \bf 0.62 & 0.52 & 0.58 & 0.52 & 0.47 & 0.52 &  \\
        \cline{2-8}
            & 15 & \bf 0.62 & 0.52 & 0.59 & 0.53 & 0.47 & 0.52 &  \\
        \whline
    \end{tabular}
\vspace{-1em}
  \label{tab:argumentidentify}
\end{table}

We set $k = 5, 10, 15$ for AC, BL and SP.
The accuracies are shown in \tabref{tab:argumentidentify}.
From Table \ref{tab:argumentidentify}, we observe that
the accuracy of AC is higher than that of BL, SP, ReVerb and SRL.
Due to its limited scale, ReVerb cannot recognize many argument
instances in the test data, and thus often labels true arguments
as negative. SRL, on the opposite side, tends to label everything
as positive because the SRL classifier is trained based
on features extracted from the context, which remains the same
even though we exchange the arguments. Thus, SRL still labels the
argument as positive.
Comparing with BL and SP, AC considers the coverage of
verb arguments, the parsing errors and overlap of concepts to
give an optimal solution with different values of $k$.
Consequently, our algorithm generates a set of concepts
which cover more precise and diversed verb argument instances.
The accuracy decreases when we use WordNet as the taxonomy
because WordNet covers 84.82\% arguments in the test data
while Probase covers 91.69\%. Since arguments that are not
covered by the taxonomy will be labeled as incorrect
by both methods, the advantage of our algorithm is reduced.

\section{Related Work}
\label{sec:related}
We first review previous work on selectional preference,
which can be seen as an alternate way of
producing abstract concepts for verb arguments, then discuss
some known work on semantic representation of verbs.

\subsection{Selectional Preference}
The most related work to our problem (AC) is
selectional preferences (SP), which
aims at computing preferences over the classes of arguments by a verb,
given the fact that some arguments are more
suitable for certain verbs than others. For example,
``drink water'' is more plausible than ``drink desk''.
While our problem defines a controllable level of abstraction
for verb arguments, selectional preference
often outputs the preference scores for all possible classes of
arguments.  Moreover, SP doesn't consider the overlap between classes,
which results in highly similar classes/concepts.

There are several approaches to computing SP.
The original {\em class-based} approaches
generalize arguments extracted from corpus to human readable
concepts using a taxonomy such as WordNet.
The most representative of such approach was proposed
by Resnik~\shortcite{resnik1996selectional},
which is used as a comparison in this paper.
Instead of WordNet,
Pantel et al.\shortcite{pantel2003clustering}
proposed a clustering method (named CBC) to
automatically generate semantic classes, which are nonetheless
not human readable.
Another recent piece of work about SP from
Fei et al.\shortcite{fei2015illinois}
doesn't abstract the arguments, and is thus different from our approach.
Other approaches including {\em cut-based} SP~\cite{li1998generalizing},
{\em similarity-based} SP~\cite{clark2001class,erk2007simple},
and {\em generative model-based} SP~\cite{ritter2010latent}
are less relevant to our problem,
because they cannot generate human readable classes.

\subsection{Semantic Representation of Verbs}
From past literature, the semantics of a word (including verbs) can be
represented by the {\em context} it appears in~\cite{mikolov2013efficient,mikolov2013distributed,mikolov2013linguistic,levy2014dependencybased}.
There are a number of ways to define the context. The simpliest is by the
words from a surrounding window.
A slightly different type of context takes advantage of structural information
in the corpus, e.g., Wikipedia. The third type of context comes from
a knowledge base or lexicon,
such as WordNet. For a verb, the gloss, its hypernyms, hyponyms, antonyms and
synonyms can be used as its context~\cite{meyer2012exhibit,yang2006verb}.
Finally, most recently, the dependency structure surrounding the verb in a
sentence has been used as its context~\cite{levy2014dependencybased}.
This is also the approach adopted in this paper.

With different types of context, a common way to represent a verb is by
a vector of distributional properties, extracted from the contexts within
a large corpus. For example, LSA \cite{deerwester1990indexing}
uses the window of words as context,
while ESA \cite{gabrilovich2007computing} uses the TF-IDF score of the word w.r.t.
the article it appears in to form a vector of Wikipedia concepts.
Another popular approach is to map the word distribution in the context
into another high-dimensional space, which is known as word embedding~\cite{mikolov2013distributed}.
Our approach can be thought of as mapping the words in the context, in this
case, the subject and object arguments into a hierarchical concept space.

\section{Conclusion}
\label{sec:conclude}
We developed a data-driven approach that automatically
infers a set of argument concepts for a verb by abstracting from a large
number of argument instances parsed from raw text.
These argument concepts are human-readable and machine-computable,
and can be used to represent the meaning of the verb.
Our evaluation demonstrates that the concepts inferred are
accurate to the human judges and show good potential at
identifying correct arguments for verbs even though such
arguments have never been seen before. This work can also
be seen as mining predicate relations between abstract noun
concepts from a taxonomy. As future work, one may consider
other important NLP tasks such as word sense disambiguation or
term similarity using the argument concept representation of
verbs.

\section*{Acknowledgments}
Kenny Q. Zhu is the corresponding author, and is partially supported by
the 2013 Google Faculty Research Award. Zhiyuan Cai and Youer Pu
contributed to the earlier version of this paper. We thank
Dr. Haixun Wang and the anonymous reviewers for their valuable comments.
\bibliographystyle{aaai}
\bibliography{ac}

\begin{thebibliography}{}

\bibitem[\protect\citeauthoryear{Abend, Reichart, and
  Rappoport}{2009}]{abend2009unsupervised}
Abend, O.; Reichart, R.; and Rappoport, A.
\newblock 2009.
\newblock Unsupervised argument identification for semantic role labeling.
\newblock In {\em ACL/IJCNLP'09},  28--36.
\newblock Association for Computational Linguistics.

\bibitem[\protect\citeauthoryear{Baker, Fillmore, and
  Lowe}{1998}]{baker1998berkeley}
Baker, C.~F.; Fillmore, C.~J.; and Lowe, J.~B.
\newblock 1998.
\newblock The berkeley framenet project.
\newblock In {\em Proceedings of the 17th international conference on
  Computational linguistics-Volume 1},  86--90.
\newblock Association for Computational Linguistics.

\bibitem[\protect\citeauthoryear{{Cambridge Dictionaries
  Online}}{2015}]{cambridge}
{Cambridge Dictionaries Online}.
\newblock 2015.
\newblock eat definition, meaning - what is eat in the {Cambridge British
  English Dictionary}.
\newblock \url{http://dictionary.cambridge.org/dictionary/british/eat}.

\bibitem[\protect\citeauthoryear{Chen \bgroup et al\mbox.\egroup
  }{2010}]{chen2010semafor}
Chen, D.; Schneider, N.; Das, D.; and Smith, N.~A.
\newblock 2010.
\newblock {SEMAFOR}: Frame argument resolution with log-linear models.
\newblock In {\em Proceedings of the 5th international workshop on semantic
  evaluation},  264--267.
\newblock Association for Computational Linguistics.

\bibitem[\protect\citeauthoryear{Clark and Weir}{2001}]{clark2001class}
Clark, S., and Weir, D.
\newblock 2001.
\newblock Class-based probability estimation using a semantic hierarchy.
\newblock In {\em NAACL'01},  1--8.
\newblock Association for Computational Linguistics.

\bibitem[\protect\citeauthoryear{Deerwester \bgroup et al\mbox.\egroup
  }{1990}]{deerwester1990indexing}
Deerwester, S.~C.; Dumais, S.~T.; Landauer, T.~K.; Furnas, G.~W.; and Harshman,
  R.~A.
\newblock 1990.
\newblock Indexing by latent semantic analysis.
\newblock {\em JAsIs} 41(6):391--407.

\bibitem[\protect\citeauthoryear{Erk}{2007}]{erk2007simple}
Erk, K.
\newblock 2007.
\newblock A simple, similarity-based model for selectional preferences.
\newblock In {\em ACL'07},  216--223.

\bibitem[\protect\citeauthoryear{Fader, Soderland, and
  Etzioni}{2011}]{fader2011identifying}
Fader, A.; Soderland, S.; and Etzioni, O.
\newblock 2011.
\newblock Identifying relations for open information extraction.
\newblock In {\em EMNLP'11},  1535--1545.
\newblock Association for Computational Linguistics.

\bibitem[\protect\citeauthoryear{Fei \bgroup et al\mbox.\egroup
  }{2015}]{fei2015illinois}
Fei, Z.; Khashabi, D.; Peng, H.; Wu, H.; and Roth, D.
\newblock 2015.
\newblock Illinois-profiler: Knowledge schemas at scale.

\bibitem[\protect\citeauthoryear{Gabrilovich and
  Markovitch}{2007}]{gabrilovich2007computing}
Gabrilovich, E., and Markovitch, S.
\newblock 2007.
\newblock Computing semantic relatedness using wikipedia-based explicit
  semantic analysis.
\newblock In {\em IJCAI}, volume~7,  1606--1611.

\bibitem[\protect\citeauthoryear{Gildea and Palmer}{2002}]{gildea2002necessity}
Gildea, D., and Palmer, M.
\newblock 2002.
\newblock The necessity of parsing for predicate argument recognition.
\newblock In {\em ACL'02},  239--246.
\newblock Association for Computational Linguistics.

\bibitem[\protect\citeauthoryear{Goldberg and
  Orwant}{2013}]{goldberg2013dataset}
Goldberg, Y., and Orwant, J.
\newblock 2013.
\newblock A dataset of syntactic-ngrams over time from a very large corpus of
  english books.
\newblock In {\em Second Joint Conference on Lexical and Computational
  Semantics (* SEM)}, volume~1,  241--247.

\bibitem[\protect\citeauthoryear{Google}{2013}]{googlengram}
Google.
\newblock 2013.
\newblock Google syntactic n-gram.
\newblock
  \url{http://commondatastorage.googleapis.com/books/syntactic-ngrams/index.html}.

\bibitem[\protect\citeauthoryear{Harris}{1954}]{harris1954distributional}
Harris, Z.~S.
\newblock 1954.
\newblock Distributional structure.
\newblock {\em Word}.

\bibitem[\protect\citeauthoryear{Kipper \bgroup et al\mbox.\egroup
  }{2000}]{kipper2000class}
Kipper, K.; Dang, H.~T.; Palmer, M.; et~al.
\newblock 2000.
\newblock Class-based construction of a verb lexicon.
\newblock In {\em AAAI/IAAI},  691--696.

\bibitem[\protect\citeauthoryear{Levy and
  Goldberg}{2014}]{levy2014dependencybased}
Levy, O., and Goldberg, Y.
\newblock 2014.
\newblock Dependencybased word embeddings.
\newblock In {\em ACL'14}, volume~2,  302--308.

\bibitem[\protect\citeauthoryear{Li and Abe}{1998}]{li1998generalizing}
Li, H., and Abe, N.
\newblock 1998.
\newblock Generalizing case frames using a thesaurus and the {MDL} principle.
\newblock {\em Computational linguistics} 24(2):217--244.

\bibitem[\protect\citeauthoryear{Manning \bgroup et al\mbox.\egroup
  }{2014}]{manning2014stanford}
Manning, C.~D.; Surdeanu, M.; Bauer, J.; Finkel, J.; Bethard, S.~J.; and
  McClosky, D.
\newblock 2014.
\newblock The {Stanford} {CoreNLP} natural language processing toolkit.
\newblock In {\em ACL'14},  55--60.

\bibitem[\protect\citeauthoryear{Meyer and Gurevych}{2012}]{meyer2012exhibit}
Meyer, C.~M., and Gurevych, I.
\newblock 2012.
\newblock To exhibit is not to loiter: A multilingual, sense-disambiguated
  wiktionary for measuring verb similarity.
\newblock In {\em COLING},  1763--1780.

\bibitem[\protect\citeauthoryear{Meza-Ruiz and Riedel}{2009}]{meza2009jointly}
Meza-Ruiz, I., and Riedel, S.
\newblock 2009.
\newblock Jointly identifying predicates, arguments and senses using markov
  logic.
\newblock In {\em NAACL'09},  155--163.
\newblock Association for Computational Linguistics.

\bibitem[\protect\citeauthoryear{Mikolov \bgroup et al\mbox.\egroup
  }{2013a}]{mikolov2013efficient}
Mikolov, T.; Chen, K.; Corrado, G.; and Dean, J.
\newblock 2013a.
\newblock Efficient estimation of word representations in vector space.
\newblock {\em arXiv preprint arXiv:1301.3781}.

\bibitem[\protect\citeauthoryear{Mikolov \bgroup et al\mbox.\egroup
  }{2013b}]{mikolov2013distributed}
Mikolov, T.; Sutskever, I.; Chen, K.; Corrado, G.~S.; and Dean, J.
\newblock 2013b.
\newblock Distributed representations of words and phrases and their
  compositionality.
\newblock In {\em NIPS'13},  3111--3119.

\bibitem[\protect\citeauthoryear{Mikolov, Yih, and
  Zweig}{2013}]{mikolov2013linguistic}
Mikolov, T.; Yih, W.-t.; and Zweig, G.
\newblock 2013.
\newblock Linguistic regularities in continuous space word representations.
\newblock In {\em HLT-NAACL},  746--751.

\bibitem[\protect\citeauthoryear{Miller and
  Charles}{1991}]{miller1991contextual}
Miller, G.~A., and Charles, W.~G.
\newblock 1991.
\newblock Contextual correlates of semantic similarity.
\newblock {\em Language and cognitive processes} 6(1):1--28.

\bibitem[\protect\citeauthoryear{Miller and Fellbaum}{1998}]{miller1998wordnet}
Miller, G., and Fellbaum, C.
\newblock 1998.
\newblock Wordnet: An electronic lexical database.

\bibitem[\protect\citeauthoryear{{Oxford University Press}}{2015}]{oxford}
{Oxford University Press}.
\newblock 2015.
\newblock eat - definition of eat in {English} from the {Oxford} dictionary.
\newblock \url{http://www.oxforddictionaries.com/definition/english/eat}.

\bibitem[\protect\citeauthoryear{Pantel}{2003}]{pantel2003clustering}
Pantel, P.~A.
\newblock 2003.
\newblock {\em Clustering by committee}.
\newblock Ph.D. Dissertation, Citeseer.

\bibitem[\protect\citeauthoryear{Resnik}{1996}]{resnik1996selectional}
Resnik, P.
\newblock 1996.
\newblock Selectional constraints: An information-theoretic model and its
  computational realization.
\newblock {\em Cognition} 61(1):127--159.

\bibitem[\protect\citeauthoryear{Ritter, Etzioni, and
  others}{2010}]{ritter2010latent}
Ritter, A.; Etzioni, O.; et~al.
\newblock 2010.
\newblock A latent dirichlet allocation method for selectional preferences.
\newblock In {\em ACL'10},  424--434.
\newblock Association for Computational Linguistics.

\bibitem[\protect\citeauthoryear{Wu \bgroup et al\mbox.\egroup
  }{2012}]{wu2012probase}
Wu, W.; Li, H.; Wang, H.; and Zhu, K.~Q.
\newblock 2012.
\newblock Probase: A probabilistic taxonomy for text understanding.
\newblock In {\em SIGMOD'12},  481--492.
\newblock ACM.

\bibitem[\protect\citeauthoryear{Yang and Powers}{}]{yang2006verb}
Yang, D., and Powers, D.~M.
\newblock {\em Verb similarity on the taxonomy of WordNet}.
\newblock Citeseer.

\end{thebibliography}

\end{document}